\documentclass[review]{elsarticle}

\journal{Pattern Recognition}

\usepackage{graphicx}
\usepackage{threeparttable}
\usepackage{amsmath}
\usepackage{multirow}
\usepackage{booktabs}
\usepackage{graphicx}
\usepackage{tabularx}
\usepackage{float}
\usepackage{amsopn}
\usepackage{xcolor}
\usepackage{makecell}
\usepackage{caption}
\usepackage[ruled,norelsize,linesnumbered]{algorithm2e}
\usepackage{stfloats} 

\makeatletter
\newcommand{\removelatexerror}{\let\@latex@error\@gobble}
\makeatother

\newcommand{\bitem}{\begin{itemize}}

\newcommand{\eitem}{\end{itemize}}
\newcommand{\benum}{\begin{enumerate}}
\newcommand{\eenum}{\end{enumerate}}
\newcommand{\bdm}{\begin{displaymath}}

\newcommand{\edm}{\end{displaymath}}
\newcommand{\beq}{\begin{equation}}
\newcommand{\bea}{\begin{eqnarray}}
\newcommand{\eea}{\end{eqnarray}}

\newcommand{\barray}{\begin{displaymath} \begin{array}{rcl}}
\newcommand{\earray}{\end{array}\end{displaymath}}
\newcommand{\eeq}{\end{equation}}

\newcommand{\bfu}{{\bf u}}
\newcommand{\bfx}{{\bf x}}

\newcommand{\bfy}{{\bf y}}

\newcommand{\bfa}{{\bf a}}

\newcommand*{\Scale}[2][4]{\scalebox{#1}{$#2$}}%

\begin{document}

\begin{frontmatter}

\title{Kernel-based Generative Learning in Distortion Feature Space}

\author[firstaddress]{Bo Tang}
\author[secondaddress]{Paul M. Baggenstoss}
\author[firstaddress]{Haibo He\corref{mycorrespondingauthor}}
\ead{he@ele.uri.edu}
\cortext[mycorrespondingauthor]{Corresponding author}


\address[firstaddress]{Department of Electrical, Computer, and Biomedical Engineering, \\ University of Rhode Island, Kingston, RI, 02881}
\address[secondaddress]{Frauhnhofer FKIE, Fraunhoferstr 20, 53343 Wachtberg, Germany}

%
%
%

\begin{abstract}
This paper presents a novel kernel-based generative classifier which is defined in a distortion subspace using polynomial series expansion, named Kernel-Distortion (KD) classifier. An iterative kernel selection algorithm is developed to steadily improve classification performance by repeatedly removing and adding kernels. The experimental results on character recognition application not only show that the proposed generative classifier performs better than many existing classifiers, but also illustrate that it has different recognition capability compared  to the state-of-the-art discriminative classifier - deep belief network. The recognition diversity indicates that a hybrid combination of the proposed generative classifier and the discriminative classifier could further improve the classification performance. Two hybrid combination methods, cascading and stacking, have been implemented to verify the diversity and the improvement of the proposed classifier.
\end{abstract}

\begin{keyword}
Distortion feature space \sep kernel-based generative classifier \sep hybrid classification \sep deep belief nets \sep character recognition
\end{keyword}

\end{frontmatter}


\section{Introduction}

Learning and inference are two important aspects for any machine learning application. For a classification problem, while the learning process aims to obtain underlying data model given a set of training data, the inference process attempts to make a prediction decision for a test data using the learned model. There are two kinds of classifiers in the learning process for classification: \textit{generative} classifier and \textit{discriminative} classifier. While the discriminative classifier learns a mapping function from input variables to output class labels, the generative classifier learns underlying joint distribution models of the given data. Generative learning is very useful for many machine learning applications, since it provides an insight to understand the learned data structures from the learning process. 

Although many generative and discriminative classifiers have been proposed, most previous research generally makes a standard assumption that the test data should be sampled from the same distribution as the given training data. However, this assumption is inappropriate for many real world classification applications, which results in misclassifications of those unseen test data drawn from a different distribution. One of well-known examples is handwritten digits classification problem in which some unseen test images with some distortions are difficult to be modeled with the same distribution as the training data. Most of the existing standard classifiers fail to identify these test images even with slight distortions. Interestingly, these misclassifications can be easily recognized by humans, as these handwritten digits just have some translations, expansions, and rotations. To address this issue, several distortion-invariant discriminative classifiers have been proposed. For example, Jarrett et al. in \cite{jarrett2009best} presented a model for hierarchical feature extraction with several different layers. Above the filter bank layer and the non-linear transformation layer, a pooling layer is used to average filter outputs over local neighbors, which ensures the classifier is invariant to small distortions, thereby improving the classification performance. Kato et al. in \cite{kato1999handwritten} extracted directional element features to detect partial inclination and reduce undesired effects on degraded images, and Liu in \cite{liu2007normalization} extracted normalization-cooperated gradient features to alleviate the effect of stroke direction distortion, both of which can provide superior performance in character recognition. More recently, deep neural networks, e.g., deep belief nets and deep convolutional neural networks, trained with numerous artificially distorted data achieve promising performance improvement \cite{hinton2006fast}\cite{cirecsan2011handwritten}\cite{cirecsan2011convolutional}\cite{cirecsan2011flexible}. Transfer learning, or domain adaption, is the other way to improve the prediction of unseen test data using the knowledge learned from one or more source data sets \cite{pan2010survey}\cite{huang2012boosting}\cite{wang2014unsupervised}.


In this paper, we propose a new kernel density estimation in the distortion subspace using polynomial series expansion, and build a generative classification approach for character recognition. We name our novel classification method Kernel-Distortion (KD) classifier. The distortion subspace is a linear subspace in which predictable distortions of kernels are computed using matrix polynomial expansion of differential linear operators \cite{BagImDist}. The predictable distortions include translation, expansion and rotation. To find the best kernels to model the data distribution, we also develop an effective method to iteratively select kernels with an assignment probability. Experimental results demonstrate that the proposed kernel-based generative classifier outperforms many other existing generative classifiers, such as naive Bayes, mixture Gaussian classifier, and Gaussian/Laplacian kernel density estimation classifier, and the state-of-the-art discriminative classifiers, such as nearest neighbors, multilayer perceptron neural network, support vector machine, etc. We also show that it offers better recognition capability for the testing images with slight distortions which are usually misclassified by the state-of-art classifier, such as a well-trained deep belief network \cite{hinton2006fast}. This diversity implies that a hybrid combination of the two could further improve the classification accuracy. To the best use of these two different worlds (i.e., the generative classifier and the discriminative classifier), we implement cascading and stacking hybrid combination methods. The experimental results show that these two hybrid combination methods obtain higher accuracy than either the generative classifier or the discriminative classifier.

The rest of this paper is organized as follows: In Section 2, we present related work in generative classifiers and their combination and comparison with discriminative classifier. In Section 3, we introduce our proposed kernel-based generative classification method which is defined in a distortion subspace using polynomial series expansion. A kernel selection algorithm is also developed to steadily improve classification performance. In Section 4, we introduce two hybrid combination methods: cascading and stacking to take advantage of the recognition diversity of the proposed generative classifier. In Section 5, experimental results and analysis on both MNIST and USPS handwritten digits data sets are presented to demonstrate the effectiveness of our proposed methods. Finally, a discussion and conclusion are provided in Section 6.

\section{Related Work}
For pattern recognition, both generative and discriminative approaches are two well-known classification methods. The generative classification approach learns the class-wise probability distribution $p(\mathbf{x} | y)$ from the training data, where $\mathbf{x}$ is the input data vector and $y$ is the corresponding class label. The classification decision is made on the basis of the posterior probability $p(y|\mathbf{x}) \propto p(\mathbf{x} | y) p(y)$ according to the Bayesian rule. The generative classifiers assume that the distribution $p(\mathbf{x}|y)$ could be estimated using some parametric or non-parametric methods from the training data. The parametric classifiers usually assume that the class-wise distribution $p(\mathbf{x}|y)$ is known but some parameters are unknown and need to be estimated \cite{duda2012pattern} \cite{kay2016probability} \cite{tang2016EEF} \cite{tang2016toward} \cite{tang2016bayesian}, such as naive Bayes classifier with Gaussian distribution models. In nonparmetric classifiers, the most popular approach to class-wise density estimation is the kernel density estimation (KDE) \cite{rosenblatt1956remarks} \cite{parzen1962estimation} \cite{scott2009multivariate} \cite{simonoff1996smoothing}. The non-parametric multivariate density estimation approaches offer a greater flexibility in modeling a given dataset, and have been successfully applied in applications like classification \cite{simonoff1996smoothing} \cite{xie1993vector} \cite{tang2015parametric}, discriminative inference \cite{memisevic2012shared}, background modeling \cite{patwardhan2008robust}, deformable shape and appearance modeling \cite{huang2008metamorphs} and object tracking in video analysis \cite{comaniciu2003kernel}. Our proposed generative classifier falls into this category, but estimates the density distribution in a distortion subspace to improve character recognition performance. Traditionally, the kernel density estimation constructs the density distribution by locating a kernel, usually a Gaussian kernel, at each observed data with an either fixed bandwidth or variable bandwidth. However, it is well known that most kernels do not use any prior knowledge that we may have about the data and usually suffer from boundary bias \cite{marron1994transformations} \cite{botev2010kernel}. Given the application of character recognition, our proposed Kernel-Distortion classifier takes the distortion information into account and selects the optimal kernels for density estimate. Unlike the existing kernel density estimation methods, the bandwidth of each kernel is determined in the distortion space.

\begin{figure*}[tp]
  \begin{center}
    \includegraphics[scale = 0.38]{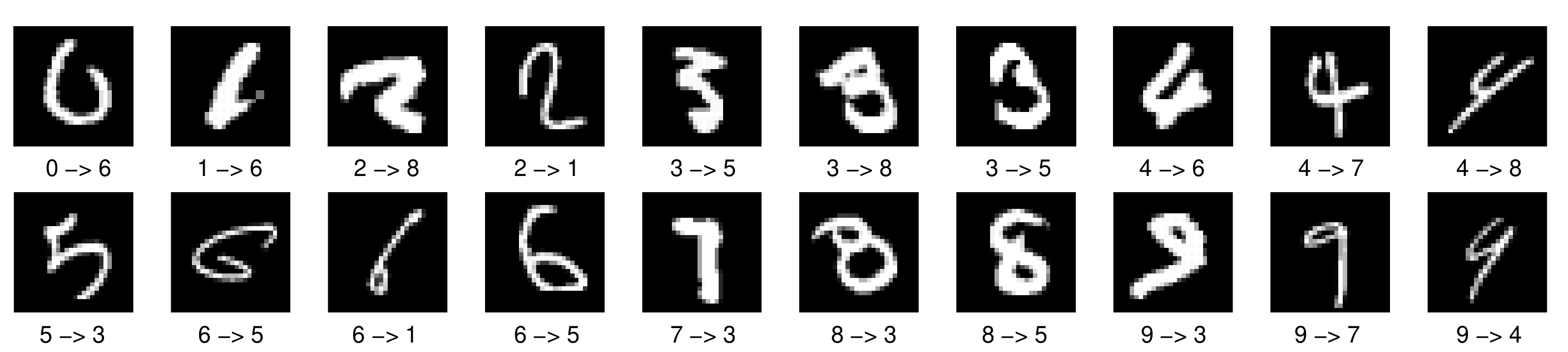}
  \caption{Several digits that have some translations, expansions and rotations from the training
  data are misclassified by a well-trained  $768 \times 500 \times 500 \times
  2000 \times 10$ deep belief network. }
  \label{example_misclassification}
  \end{center}
\end{figure*}

Contrarily, the discriminative classifiers model the posterior probability $p(y|\mathbf{x})$ directly, or learn the mapping function from input variables to output class labels from the given training data. Some well-known discriminative classifiers include neural network, nearest neighbor, and support vector machine.  There is also a wide debate about which classifier is better than the other. Many experimental results published in the literature have shown that the discriminative approach always outperforms the generative one over several real-life classification data sets \cite{rubinstein1997discriminative} \cite{nigam1999using} \cite{vapnik1998statistical} \cite{tang2015enn}. However, the answer is not
as simple as they pointed out \cite{drummond2006discriminative}. In \cite{drummond2006discriminative}, an in-depth experimental comparison between discriminative and generative classifiers was conducted, showing that the variants of generative classifiers can improve the classification performance. More comparison between these two kinds of classifiers can be seen in \cite{Mitchell97, Long07, Shai2001, ngdiscriminative, jebara02discriminative, pernkopf2005, Drummond05, Yuret2008, Schmah08RBM}. For many real-life classification problems, the discriminative classifiers usually outperform the generative classifiers, due to the fact that the assumption that training data should satisfy a specific distribution model is always inappropriate for generative classifiers, specifically for the high dimensional data. However, for the problem of character recognition, a well trained discriminative classifier still misclassifies many images that usually have some distortions from the given training images, shown in Fig. \ref{example_misclassification}. 

The difference between discriminative and generative classifiers also inspires researchers to take advantage of the two and propose combination methods into an ideal classifier \cite{raina2003classification, li2005generative, fujino2005hybrid, HolubWP08, btang2014hybrid, bosch2008scene}. One type of these combined methods is to incorporate generative models in discriminative classifiers.
Particularly, kernel-based generative models play an important role in this
combination. For example, the Fisher kernel function was combined with a support vector machine to obtain a substantial improvement in classification \cite{jaakkola1999exploiting}, and the kernel
density estimation was used to form a probabilistic neural network
which approximates the optimal Bayes rule \cite{specht1990probabilistic}. Ensemble is the other effort to make use of the diverse recognition ability of these two different classifiers. In \cite{raina2003classification}, Raina et al. proposed a hybrid model for text classification in which a large subset of the parameters were trained to maximize the generative likelihood, and a small subset of the parameters were
discriminatively trained to maximize the conditional likelihood. In \cite{li2005generative}, Li et al. presented two phases learning, a generative phase followed by a discriminative phase for object recognition in outdoor scenes, also showing a significant improvement in image retrieval. In this paper, we first propose a new generative classifier that shows a diverse recognition capacity compared to a well-trained discriminative classifier. Then, we implement two hybrid combination methods, cascading and stacking, to verify the diversity of the proposed generative classifier which can further improve the prediction performance.


%

\section{Proposed Generative Classifier}
\subsection{Model Description}
Considering an $M$-class classification problem, 
we classify a sample $\bfx$ by maximizing the {\it a posteriori} class probability, i.e.,
\begin{align}
\label{MAP_Likelihood}
\hat{i} & = \arg \max_{m=1}^M \; p(H_m|\bfx) \nonumber\\
& = \arg \max_{m=1}^M \; p(\bfx | H_m) \; p(H_m)
\end{align}
where $H_m$ denotes that the sample is classified to the $m$-th class. Using the Bayes' rule, the classifier relates to the class-wise likelihood $p(\bfx | H_m)$ and prior probability $p(H_m)$. 
We write the class-wise likelihood function $p(\bfx | H_m)$ as kernel mixtures,
\begin{align}
\label{mix1}
p(\bfx| H_m) = \sum_{k=1}^K \; w_{mk} \; p_k(\bfx| H_m)
\end{align}
where $p_k(\bfx| H_m)$ are the individual kernel density distribution functions (PDFs) and $w_{mk}$ are the prior probability of kernels which have 
\begin{align}
\sum_{k=1}^K \; w_{mk}=1
\end{align}
The kernel PDF $p_k(\bfx| H_m)$ is centered at a chosen training character $\bfx_{mk}$
and the distribution is defined using a multivariate Gaussian model, $p_k(\bfx| H_m) \sim
\mathcal{N}(\bfx_{mk}, \mathbf{C}_{mk})$. Unlike the existing method of Gaussian mixtures,
we determine both $\bfx_{mk}$ and $\mathbf{C}_{mk}$ using the distortion
subspace analysis.          

\subsection{Distortion Subspace Analysis}
The distortion subspace is the linear subspace in which predictable distortions
of the kernel center $\bfx_{mk}$ are contained. Predictable distortions
consist of translation, rotation, and expansion (contraction). The distortion subspace is computed for each kernel
center and can be derived using matrix polynomial expansion of differential linear operators \cite{BagImDist}.

Let $\bfx$ be a column-vector, the $N^2\times 1$ concatenated pixels of 
an $N\times N$ reference image.  Let $\bfy$ be an {\it slightly distorted}
version of $\bfx$. Let ${\bf P}$ be a linear differential
distortion operator.  In other words, $\bfy = {\bf P} \bfx$
is a slightly-distorted version of $\bfx$ in some
distortion space, such as rotation, translation, expansion. 
To achieve significant distortion, we apply the
operator $k$ times, $\bfy = {\bf P}^k \bfx$.
Since the operator ${\bf P}$ causes {\it slight} distortion,
we may write
$${\bf P} =  {\bf I} + \tilde{\bf P}, $$ where ${\bf I}$ is the identity matrix
and $\tilde{\bf P}$ is a matrix for distortion.
Let $|\tilde{\bf P}|$ be defined as a matrix norm,
such as the magnitude of the largest eigenvalue of $\tilde{\bf P}$.
We assume $$|\tilde{\bf P}| << 1.$$
approaches zero. With the Taylor series expansion, we may write that
\begin{align}
{\bf P}^k & = \left[ {\bf I} + \tilde{\bf P}\right]^k \nonumber \\
& = {\bf I} + k \tilde{\bf P} + \frac{k(k-1)}{2} \tilde{\bf P}^2 + 
\frac{k(k-1)(k-2)}{6} \tilde{\bf P}^3 + \cdots \nonumber
\end{align}
Since $|\tilde{\bf P}| << 1$, the above series can be truncated to some
power $p$.
Thus,
$$\bfy \simeq  \bfx + k \tilde{\bf P}\bfx + \frac{k(k-1)}{2} \tilde{\bf P}^2\bfx + 
\frac{k(k-1)(k-2)}{6} \tilde{\bf P}^3\bfx + \cdots.$$
Each of the matrix products $\tilde{\bf P}^i \bfx$, for $0\leq i \leq p$
can be concatenated into a $N^2\times p$ vector.
The idea behind distortion subspace analysis is to collect these
vectors into an $N^2\times p$ matrix 
Thus,
$${\bf A} = \left[  \tilde{\bf P}\bfx, \;\tilde{\bf P}^2\bfx, \; \ldots, \; \tilde{\bf P}^p\bfx\right],$$
We can then write
$$\bfy \simeq \bfx a + {\bf A} \bfa,$$
for some scalar amplitude $a$ and vector of amplitudes $\bfa$.

\begin{figure}[!ht]
  \begin{center}
    \includegraphics[height=2.3in,width=2.3in]{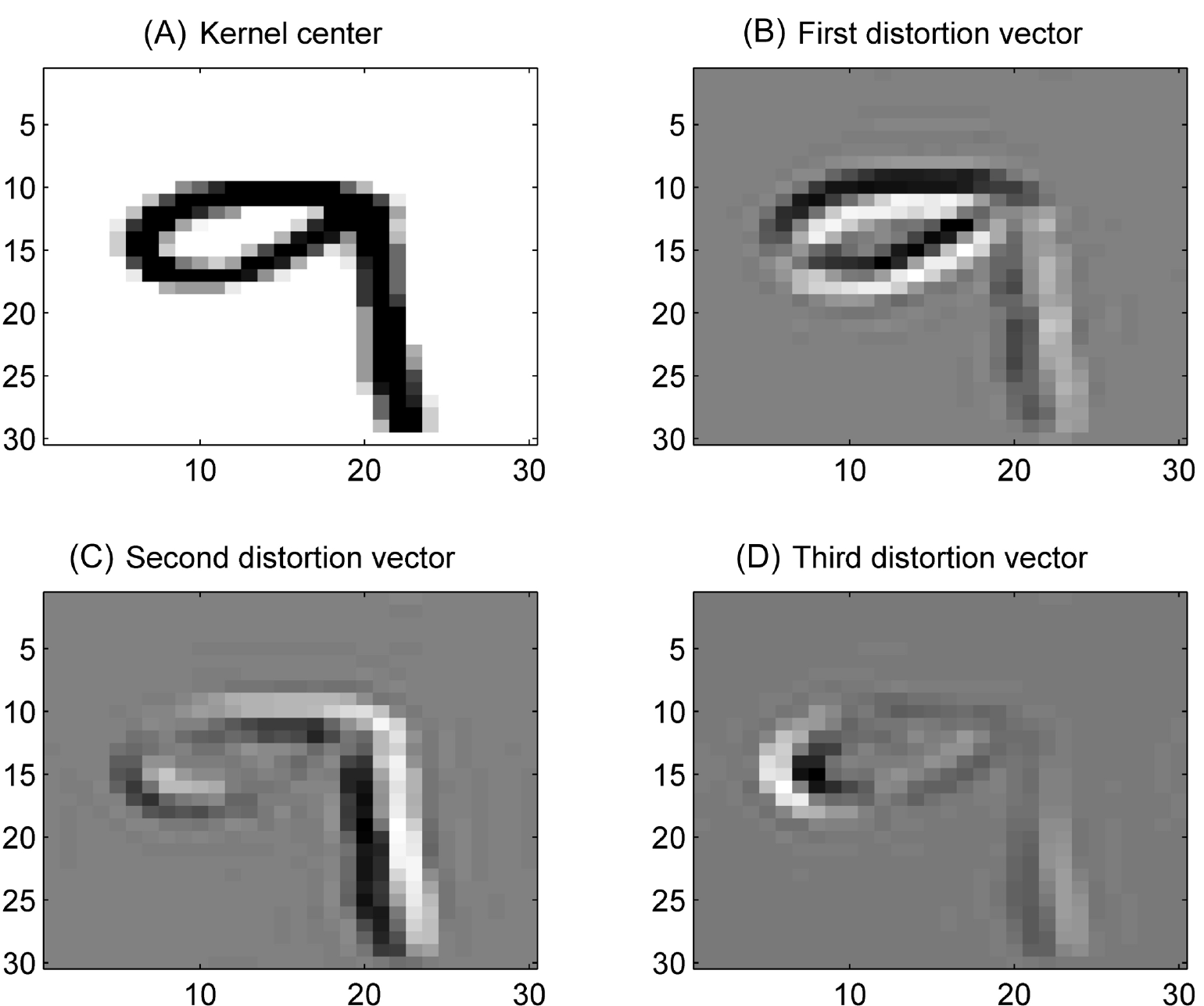}
  \caption{Example of distortion space of digit ``9": (A) Kernel center, (B) the first column vector in ${\bf U}$, (C) the second column vector in ${\bf U}$, and (D) the third column vector in ${\bf U}$.}
  \label{disto}
  \end{center}
\end{figure}

Up to now we have discussed just one distortion mode.
We will consider the following
five distortion modes:  X-translation, Y-translation, X-expansion/contraction,
Y-expansion/contraction, and rotation. Let the 
distortion operators ${\bf P}$ for the five modes
be denoted by ${\bf P}_i$ with the power of $a_i$, $\; 1\leq i \leq 5$.
In mixed distortion, we have, for example
\beq
\bfy = {\bf P}_1^{a_1} \;  {\bf P}_2^{a_2} \; {\bf P}_3^{a_3} \; \; {\bf P}_4^{a_4} \; {\bf P}_5^{a_5} \bfx.
\label{distdef}
\eeq
In this case, matrix ${\bf A}$ must contain not only the powers
of $\tilde{\bf P}_i$, but also mixed powers, with the highest total power equal to $p$, i.e., 
\begin{align}
\mathbf{A} = \left[ \tilde{\mathbf{P}}_1 \mathbf{x}, \tilde{\mathbf{P}}_2 \mathbf{x}, \cdots, \tilde{\mathbf{P}}_5 \mathbf{x}, \tilde{\mathbf{P}}_1 \tilde{\mathbf{P}}_2 \mathbf{x}, \cdots, \tilde{\mathbf{P}}^{a_1}_1 \tilde{\mathbf{P}}^{a_2}_2 \tilde{\mathbf{P}}^{a_3}_3 \tilde{\mathbf{P}}^{a_4}_4 \tilde{\mathbf{P}}^{a_5}_5\mathbf{x}, \cdots \right]
\end{align}
where $0 \leq a_j \leq p$ for $j=1, 2, \cdots, 5$ and $\sum_{j=1}^5 a_j = p$. A detailed explanation of this is given in \cite{BagImDist}, section 4.4.
Important to know is that ${\bf A}$ becomes very large.
It is therefore necessary to approximate the distortion in a simpler
way by computing the singular value decomposition (SVD) of
${\bf A}$ truncated to $q$ singular vectors,
$${\bf A} \simeq {\bf U} {\bf S} {\bf V}^T,$$
where ${\bf S}$ is the $q\times q$ diagonal matrix of the top $q$ singular values ranked in a descending order and ${\bf U}$ consists of the corresponding largest $q$ singular vectors. We then approximate $$\bfy \simeq \bfx a + {\bf U} {\bf a}.$$
The first three column vectors in ${\bf U}$ are illustrated in Fig. \ref{disto} for $\bfx$ selected from the class ``9".

\subsection{Kernel Distributions}
We use distortion subspace analysis for classification. However, we are not interested in determining the distortion powers in (\ref{distdef}). We are only interested in determining if the distortion is large, i.e. is $\|\bfa\|$ large, which can be reflected by kernel distributions. 

In our proposed Kernel-Distortion classifier, we use the following Gaussian model
\begin{align}
\label{eq:kernel_mixture}
p_k(\bfx|H_m) = \frac{1}{\sqrt{(2\pi)^{N^2} |{\bf R}_{mk} |}}
 \exp\left( \frac{ -(\bfx-\bfx_{mk})^T
{\bf R}_{mk}^{-1}  (\bfx-\bfx_{mk})}{2} \right)
\end{align}
with a special structure for ${\bf R}_{mk}$, which is given by
$${\bf R}_{mk} = \sigma^2_d \; {\bf U}_{mk} {\bf U}_{mk}^T + \sigma^2_d \tilde{\bfx}_{mk} \tilde{\bfx}_{mk}^T
  + \sigma^2_o \tilde{\bf U}_{mk} \tilde{\bf U}_{mk}^T$$
This form results from the decomposition of the energy
in $\bfx$ into three mutually orthogonal subspaces:
\benum
 \item Distortion subspace characterized by the column space of ${\bf U}_{mk}$, the $N^2\times q$ orthonormal matrix of basis functions for the {\it distortion subspace} of the kernel with center $\bfx_{mk}$.
 \item Amplitude subspace characterized by the vector $\tilde{\bfx}_{mk}$, essentially the energy in the direction of the reference image. To be exact, we have made $\bfx_{mk}$       orthogonal to ${\bf U}_{mk}$,
       $$\bfu = \bfx_{mk} - {\bf U}_{mk} {\bf U}_{mk}^T \bfx_{mk},$$
      $$\tilde{\bfx}_{mk} = {\bfu\over \|\bfu\|}.$$
 \item Noise subspace characterized by the column space of $\tilde{\bf U}_{mk}$, essentially error that cannot be explained by a scaling or distortion of the reference image. Matrix $\tilde{\bf U}_{mk}$ is the $N^2\times (N^2-q-1)$ orthogonal complement space, orthogonal to both ${\bf U}_{mk}$ and $\tilde{\bfx}_k$.
\eenum

Together, the vectors $\tilde{\bfx}_k$, ${\bf U}_k$, and $\tilde{\bf U}_k$ form a complete orthonormal basis for the space ${\cal R}^{(N^2)}$. We regard any energy in the {\it orthogonal subspace}, spanned by $\tilde{\bf U}_{mk}$, to be error and is penalized
heavier (using lower variance - $\sigma^2_o$). Any energy in either the {\it amplitude subspace}, spanned by $\tilde{\bfx}_{mk}$, or in the distortion subspace, spanned by ${\bf U}_{mk}$ is considered normal distortion and is penalized less (higher variance - $\sigma^2_d$).

\subsection{Kernel Selection}
To initialize the mixture density in Eq. (\ref{mix1}) for a given class $m$, we select $K$ random samples as kernels, then compute the distortion subspace ${\bf U}_{mk}$ with 
respect to each kernel $k$ needed to compute the kernel density $p_k(\bfx| H_m)$.
Kernel selection can be accomplished by repeatedly removing and adding kernels, always having $K$ kernels after each iteration. In each iteration, we compute the weights $w_{mk}$. Assume that there are $N_m$ training data from class $m$, and we compute a $L_m \times K$ likelihood matrix ${\bf W}$ in which each element $W_{i,k}$ is calculated as
\begin{align}
\label{eq:kernel_selection_W}
{W}_{i,k}= p_k(\bfx| H_m), \; 1\leq i \leq L_m, \;\; 1\leq k \leq K
\end{align}
Let $\tilde{\bf W}$ be the normalized version of ${\bf W}$, where
\begin{align}
\label{eq:kernel_selection_WNorm}
\sum_{k=1}^K \; \tilde{W}_{i,k} = 1
\end{align}
We define a quantity 
\begin{align}
\label{eq:kernel_selection_alpha}
\alpha_k = \sum_{i=1}^{L_m} \; \tilde{W}_{i,k}
\end{align}
The element $\tilde{W}_{i,k}$ in $\mathbf{\tilde{W}}$ is the ``weight" of data point $\mathbf{x}_i$ for the $k$-th kernel. So the quantity $\alpha_{k}$ is the sum of the membership weights for the $k$-th kernel, which is the effective number of data points assigned to the $k$-th kernel. Thus, the estimated kernel weight in Eq. (\ref{mix1}) is given by
\begin{align}
\label{eq: kernel_selection_weights}
w_{mk} = \frac{\alpha_k}{\sum_{l=1}^K\; \alpha_l}
\end{align}

Using Eq. (\ref{eq:kernel_mixture}), we compute the total log-likelihood,
\begin{align}
\label{tlik1}
Q_m=\sum_{i=1}^{L_m} \; \log p(\bfx_i|H_m).
\end{align}

We then re-compute Eq. (\ref{tlik1}) by iteratively removing each of the $K$ kernels. When a kernel is assumed to be removed, we re-normalize
the weights $w_{mk}$ so they are always summed to $1$.
The kernel that produces the least drop in $Q_m$ is deemed the
most ``expendable" kernel, and hence this kernel is removed.

Next, we identify the training sample that is most likely to increase $Q_m$ if it is added as a new kernel. To identify this sample, we select the $L_m^\prime=L_m - K$ training samples that are not already kernels. We then form the $L_m^\prime\times L_m$ Euclidean distance matrix $D_{k,l} = |\bfx_{mk} - \bfx_{ml}|,$ setting $D_{k,l}$ to infinity if $i=j$. For each sample pair $(k,l)$ selected from this set, we form the assignment probability from sample $k$ to sample $l$: $$p_{k,l} = \exp \left(-\frac{D_{k,l}}{C}\right),$$
then normalize so that $$\sum_{l=1}^{L_m} \; p_{k,l}=1.$$
After normalization, $p_{k,l}$ can be thought of the probability that sample $k$ is assigned to sample $l$. The ``value" of sample $l$ as a new kernel is then measured by the number of samples assigned to it, or by the weight $a_l$:
\begin{align}
\label{eq:kernel_selection_al}
a_l = \sum_{k=1}^{L_m^{'}} \; p_{k,l}
\end{align} 
The sample (not already used as a kernel) with the largest value $a_l$ is selected as the kernel to replace the one removed above. After repeating this process of removing and adding kernels, the likelihood is steadily increasing. This iterative process is terminated when it runs a specific number of iterations or the likelihood exceeds a predefined threshold. 

%

Let the total number of features of $\mathbf{x}$ be $D = N \times N$, and assume that the complexity of calculating the Gaussian likelihood in Eq. (\ref{eq:kernel_mixture}) is $O(D)$. Then, the computational complexity of each iteration is $O(L_m K D)$ for the $m$-th class. Hence, the total computational complexity for our kernel selection algorithm is $O(\sum_{m=1}^M L_m K D \bar{I}) = O(L_t K D \bar{I})$, where $L_t$ denotes the total number of training data and $\bar{I}$ denotes the average number of iterations over all classes. The number of iterations for each class usually depends on the size of training data set, the number of kernels, and how good the kernels are chosen initially. For the MNIST data set, when we randomly initialize $K$ kernel centers, our experimental results indicate that $500$ iterations would achieve nice classification performance as shown in Fig. \ref{iter}.

\section{Hybrid Classification}
The proposed kernel-based generative classifier is defined in a distortion subspace using the polynomial series expansion, which can offer a different classification capacity compared to the existing discriminative classifiers. A hybrid combination method can be used to take advantage of the best of both two worlds. Although there are many combination techniques in literature, only the hybrid one built with different kind of classifiers is suitable for our purpose, such as cascading and stacking methods, as we introduce as follows.      

\subsection{Cascading Method}
In our first hybrid classification method, we cascade the discriminative classifier and generative classifier with a threshold $\tau$. The idea of cascading classifiers was firstly proposed in \cite{alpaydin1998cascading} which can provide competitive performance compared with voting and boosting methods \cite{alimoglu2001combining}. Given a test digit image $\mathbf{x}$ to be classified, we input it to the discriminative classifier and obtain the posterior probability $p(H_i | \mathbf{x})$ for each class $i$. Usually, we make a classification decision by assigning $\mathbf{x}$ to the class $\hat{i}$ which has maximum posterior probability over all classes in Eq. (\ref{MAP_Likelihood}). In our cascading combination classifier, we firstly compare the maximum posterior probability with the threshold $\tau$. If it exceeds the threshold, we consider this classification result as final decision with high confidence. Otherwise, we further take it as input in the generative classifier to make final classification decision. This cascading combination method can be written as follows                          
\begin{align}
\Scale[0.9]{
\label{cascading}
\hat{i}_c = \left \{ 
\begin{array}{l l}
\arg \max_{m=1}^M \; p_{\text{d}}(H_m|\bfx) & \text{if }  \max_{m=1}^M p_{\text{d}}(H_m|\bfx) > \tau \\
\arg \max_{m=1}^M \; p_{\text{g}}(H_m|\bfx) & \text{otherwise}
\end{array}
\right.}
\end{align}
where $p_{\text{d}}(H_m|\bfx)$ and $p_{\text{g}}(H_m|\bfx)$ denote the posterior probability of the discriminative classifier and the generative classifier, respectively. Notice that if $\tau = 1$, the cascading classifier is only decided by the generative classifier, and if $\tau = 0$, it is only decided by the discriminative classier. A suitable threshold would obtain the best performance for this kind of hybrid combination classifier.

\subsection{Stacking Method}
Our second hybrid classifier is based on a voting or weighting method, also called stacking \cite{wolpert1992stacked}. We use it as a hybrid combination method because it can take a linear combination of the discriminative and generative classifiers. Compared to Bayes Model Averaging (BMA) \cite{hoeting1999bayesian}, stacking method is a non-Bayes form of model averaging where its weights are no longer posterior probabilities of averaged models and are learned from training data directly. 

In general, the stacking method linearly combines $L$ classifiers and decides output class label according to
\begin{align}
\label{voting}
\hat{i}_s = \arg \max_{m=1}^M \; \sum_l^L w_l p_{l}(H_m|\bfx)
\end{align} 
with the constraints
\begin{align}
\forall l, w_l \geq 0 \text{ and } \sum_{l}^L w_l = 1
\end{align}
where $w_l$ denotes the voting weight of the $l$-th classifier. 

Due to the diversity of recognition ability between the generative classifier and the discriminative classifiers, the consensus of them would lead to a more powerful classifier. After both learners are trained individually, the output of the final hybrid classifier is determined by
\begin{align}
\label{voting}
\hat{i}_s = \arg \max_{m=1}^M \; \left[ w p_{\text{d}}(H_m|\bfx) +
(1-w) p_{\text{g}}(H_m|\bfx) \right]
\end{align} 
where $w \in [0,1]$ is the only one parameter needed to be determined. 

We comment that other averaging approaches also exist in literature, such as functional aggregation, boosting and bagging. Even though they may also obtain high predication accuracy, we only focus on cascading and stacking methods for two reasons: The first reason is that these two methods have less computational cost compared to the methods of boosting and bagging in which a large collection of weak classifier rules are trained to build a final classification rule; The second one is that the goal of the hybrid combination methods is to show the recognition diversity and benefit of our proposed generative classifier in the distortion space. Any other hybrid combination schemes would be still consistent with our proposed generative classifier.

\section{Experimental Results and Analysis}
\subsection{Data Sets}
We use both MNIST \cite{lecun1998gradient} and USPS \cite{hull1994database}
handwritten digits data sets as our benchmarks to evaluate the performance of our proposed generative classifier and its combination with discriminative classifiers. Both of these data sets are widely used in machine learning as real-world applications. The MNIST data set of handwritten digits contains a training set of $60,000$ images, and a
test set of $10,000$ images. It has a total $10$ digits classes, ranging from $0-9$,
and contains approximately $6000$ training samples and $1000$ testing samples per class. All images have been centered and translated with a size of  $28 \times 28$. 
The USPS data set of handwritten digits contains $11,000$ images in total for $10$ classes ($0-9$ digits). All images have the size of $16 \times 16$ with grey level pixels, which have been scaled to the range of $[0 \ 1]$. In our experiments, we randomly choose $9,900$ images as training data set and the remaining $1,100$ images as test data set. In order to account for distortion, we added a 1-pixel margin, increasing the image size to $30 \times 30$ for the MNIST data set and $18 \times 18$ for the USPS data set.

\subsection{Experimental Results}

For both MNIST and USPS data sets, we compare our Kernel-Distortion classifier with other five generative classifiers, including naive Bayes classifier, linear discriminant analysis (LDA), Gaussian mixture model (GMM) classifier, Gaussian kernel density estimation (GKDE) classifier, and Laplacian kernel density estimation (LKDE) classifier, and four classic discriminative classifiers, including multilayer percepton (MLP) neural network (MLP-NN), deep belief nets (DBN), support vector machine (SVM), and $k$-nearest neighbors ($k$-NN). For both MNIST and USPS data sets, we use the same parameters settings. To avoid the issue of singular covariance matrix and to obtain better classification performance in naive Bayes, GMM and LDA classifiers, we use principle component analysis (PCA) method to reduce the number of feature dimension to $50$. The naive Bayes classifier assumes that the features are independent and each of features satisfies a single Gaussian distribution. In GMM classifier, the distribution of each digit class is modeled by $100$ mixture Gaussian components, and the Expectation-Maximization (EM) algorithm is employed to estimate the distribution model parameters including component prior probability, mean vector and covariance matrix of each Gaussian component. For both Gaussian and Laplacian KDE classifiers with the kernel width of $0.1$, we use the same $100$ kernels as our proposed approach. For all of these five generative classifiers, the maximum a posteriori (MAP) rule in Eq. (\ref{MAP_Likelihood}) is used to make classification decisions after the calculation of likelihood. 

For the discriminative classifiers, we train three-layer $784 \times 100 \times 10$ MLP neural network with one input layer ($784$ neurons), one hidden layer ($100$ neurons) and one output layer ($10$ neurons) for classification. In training stage, we set the learning rate as $0.01$, and the number of iterations as $5000$. SVM classifiers are trained with radial basis function (RBF) kernel for these two data sets. We built $10$ ``one-versus-all" SVM classifiers for the $10$-class classification problem, and assign the class label with the greatest margin to the test image. The DBN used in our experiments is a $768 \times 500 \times 500 \times 2000 \times 10$ network which is trained by two stages: pre-training stage and fine-tuning stage. In its pre-training stage, each hidden layer is trained with $50$ epochs using all training data, and in its fine-tuning stage, a back-propagation training algorithm with $200$ epochs is adopted for the discriminative purpose. We refer to the interested readers to \cite{hinton2006fast} for the detailed descriptions of this deep belief network.

\begin{figure}[!ht]
  \begin{center}
    \includegraphics[height=2.6in]{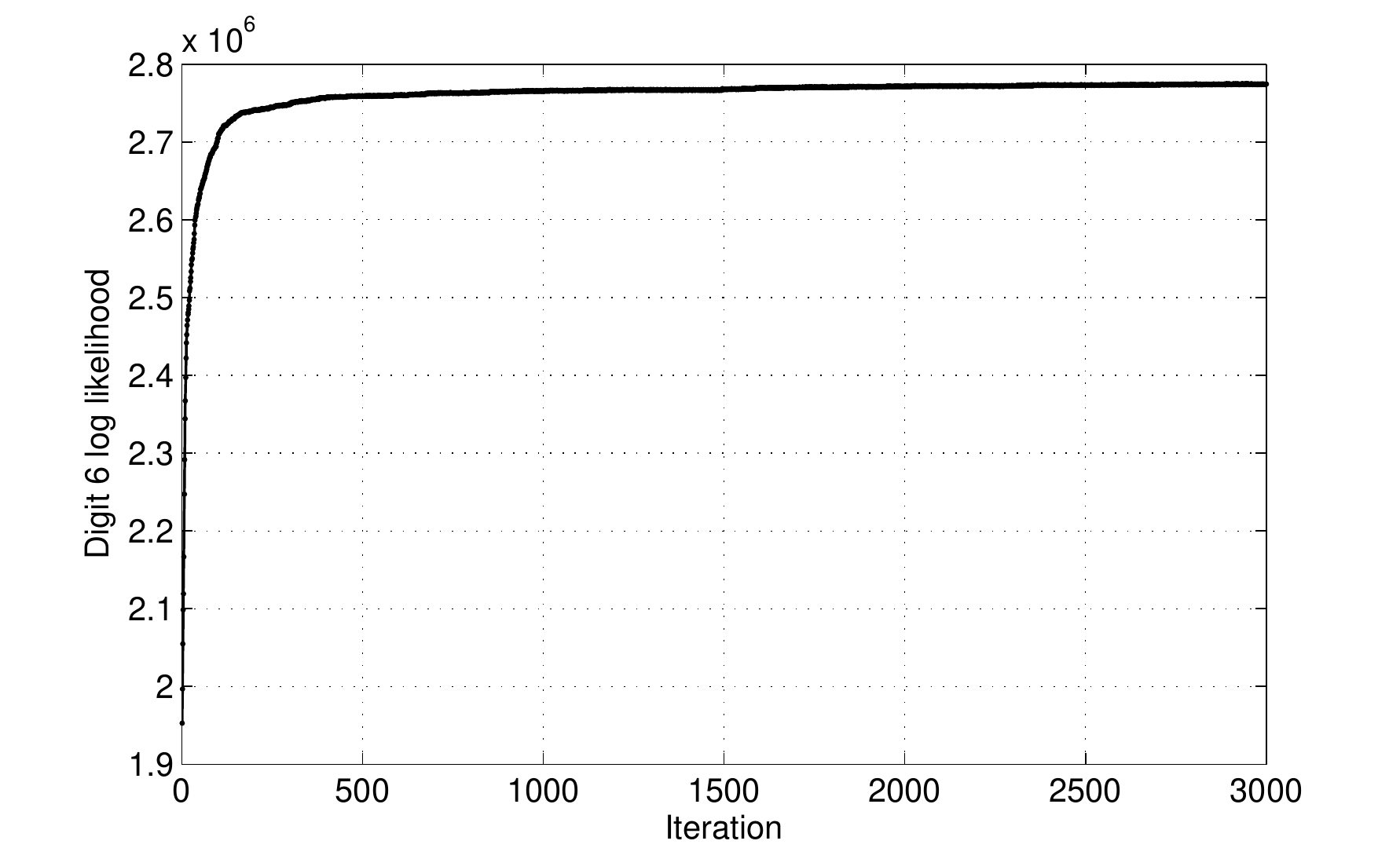}
  \caption{Training data likelihood as a function of iterations for digit ``6".  }
  \label{iter}
  \end{center}
\end{figure}


For our proposed Kernel-Distortion classifier, we use $K=100$ kernels, and set $p = 3$, $q = 40$, $\sigma_o^2 = 0.03$ and $\sigma_d^2 = 0.9$. To select better kernels, the proposed iterative kernel selection algorithm is applied. In Fig. \ref{iter}, we show the total likelihood $Q$ as a function of iteration for up to $3000$ iterations for the digit ``$6$", when the MNIST data set is used as an example. Although not purely monotonic, it shows a steadily increasing likelihood \footnote{A video demonstration for the process of kernel selection is provided as Supplementary Material. One can see that the class-wise likelihood is steadily increased with the number of iteration, and that the ``best to remove" digits seem really bad at first, and then at the end, they look almost the same as the ``best to add". }. The steady increasing likelihood demonstrates the effectiveness of our proposed iterative kernel selection algorithm. We stop the kernel-selection algorithm at the $500$-th iteration for both MNIST and USPS data sets, although the likelihood still increases. Table \ref{tab_all} shows the comparison results in terms of overall testing classification error rate. It can be shown that the discriminative classifiers usually outperform the generative classifiers for these two data sets. However, our proposed generative classifier can greatly improve the prediction performance compared with other generative classifiers and outperform some other well-trained discriminative classifiers. It performs best for USPS handwritten digits classification and ranks third for MNIST handwritten digits classification, which demonstrate the effectiveness of our proposed Kernel-Distortion classifier. 

\begin{table*}[tp]
\centering
\renewcommand {\arraystretch}{1.5}
\caption{Total testing classification error rate in percentage compared with naive Bayes,
LDA, GMM, GKDE, LKDE, MLP-NN, DBN, RBF-SVM, $k$-NN ($k=1$), and our Kernel-Distortion classifier. For each dataset, we highlight the best result with \underline{\textbf{Bold}} value, the second one with \textbf{Bold} value, and the third one with \textit{Italic} value, among all 11 classifiers.}
\label{tab_all}
\centering
\setlength{\tabcolsep}{2.0pt}
\resizebox{12cm}{!}{
{\small
\begin{threeparttable}
\begin{tabular}{c | c c c c c c c c c c}
\toprule
Data Set & Kernel-Distortion &  Naive Bayes & LDA & GMM & GKDE & LKDE & MLP-NN & DBN & RBF-SVM &
$1$-NN
\\
\hline
MNIST & $\textit{2.38} \%$ &  $13.73 \%$ & $3.67\%$ & $2.58\%$ & $6.05\%$ & $7.24\%$ &
$2.48 \%$ & $\underline{\textbf{1.08}}\%$ & $\textbf{1.74}\%$ & $3.12\%$
\\
USPS & $\underline{\textbf{1.45}}\%$ &  $10.45\%$ & $\textit{2.73}\%$ & $3.82 \%$ & $6.09\%$ & $2.64\%$ & $2.91 \%$ & $3.09\%$ &
$\textbf{2.27}\%$ & $3.55\%$
\\
\bottomrule
\end{tabular}
\end{threeparttable}}}
\end{table*}

\begin{figure}[!ht]
  \begin{center}
    \includegraphics[height=2.5in]{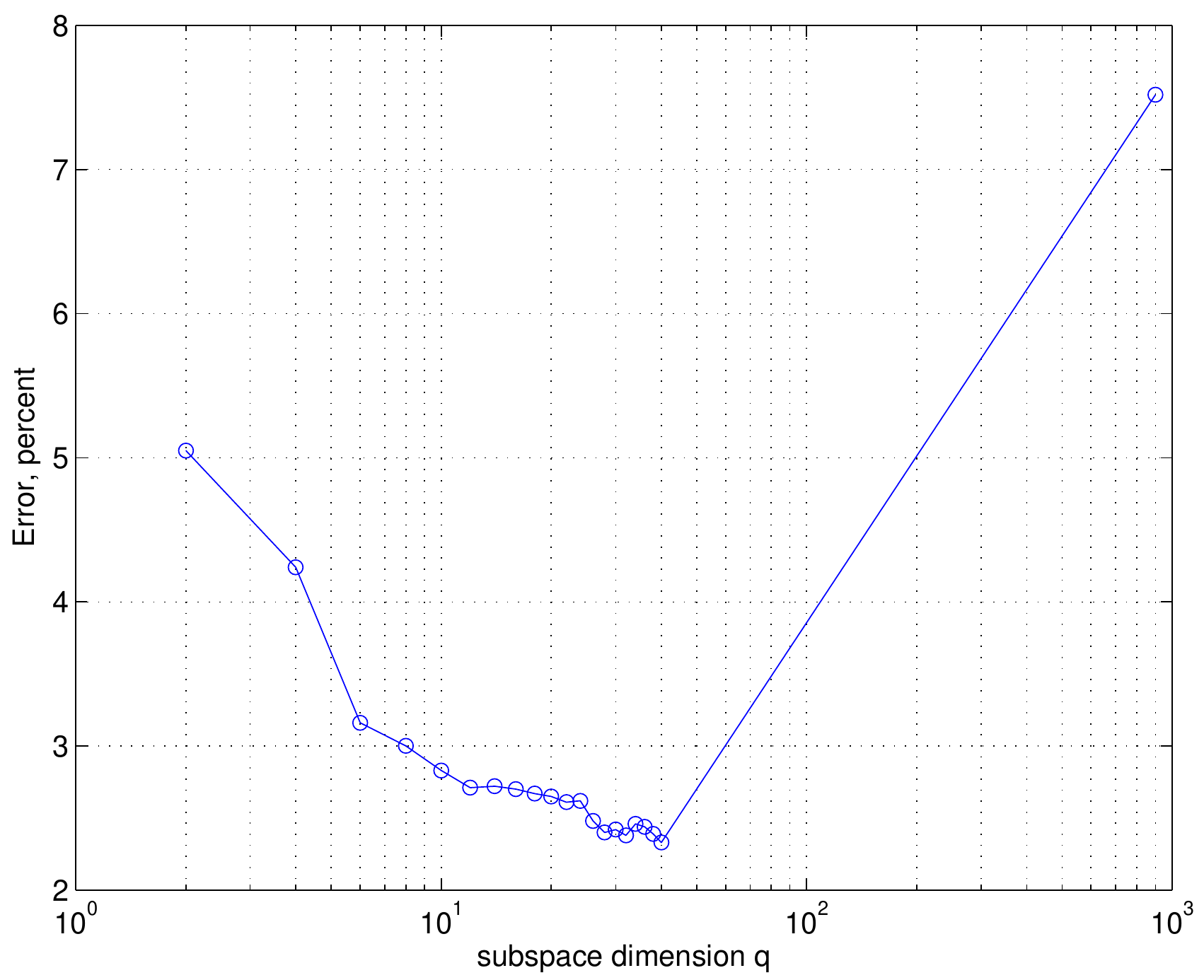}
  \caption{Total classification error as a function of $q$.
The largest value of $q$ tried was 40. The performance
at the far right is the Euclidean distance metric classifier,
which is equivalent to $q=N^2$.}
  \label{qerr}
  \end{center}
\end{figure}

\begin{figure}[!ht]
  \begin{center}
    \includegraphics[height=2.3in]{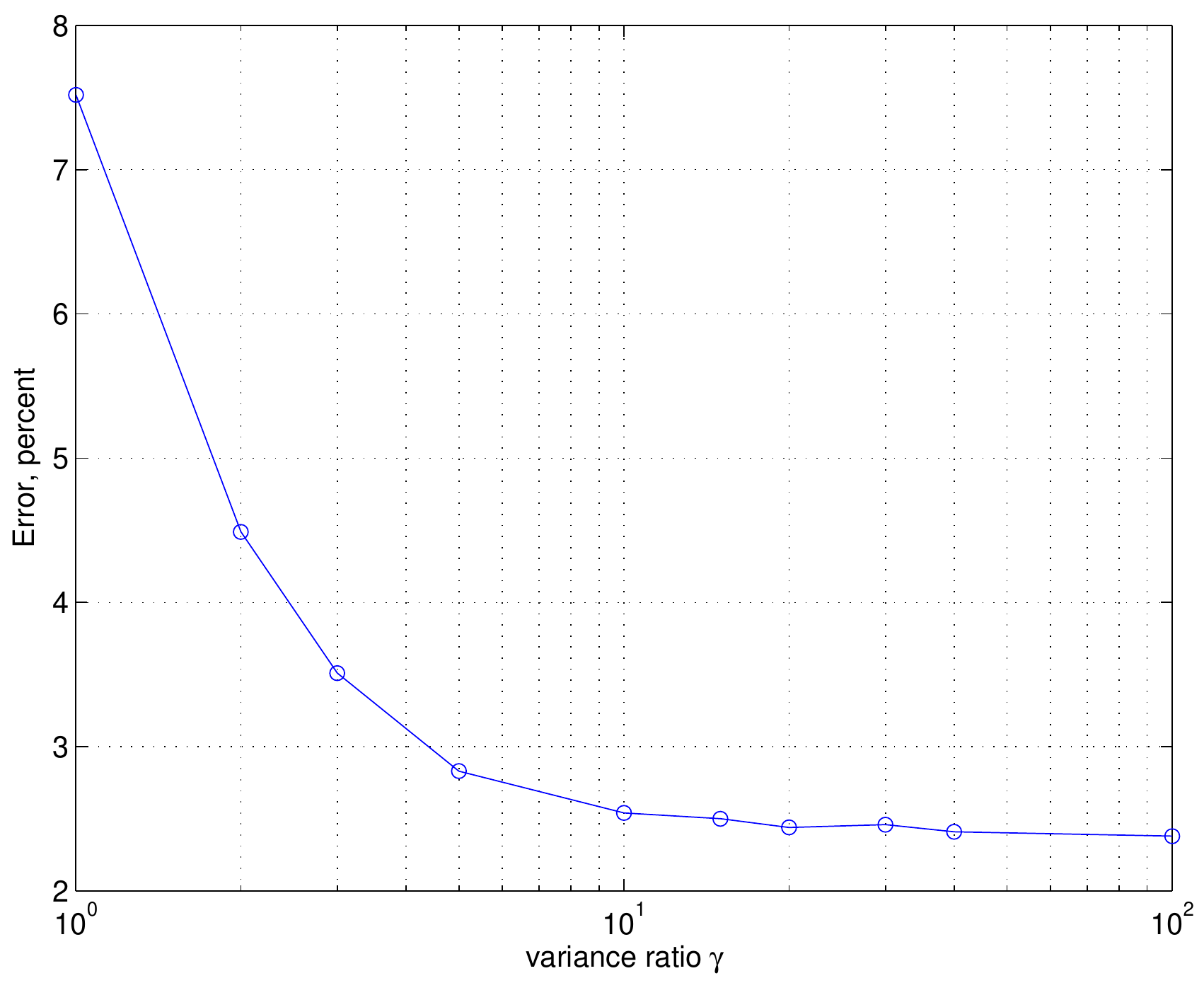}
  \caption{Total classification error as a function of
variance ratio $\sigma^2_d/\sigma^2_o$.}
  \label{verr}
  \end{center}
\end{figure}

\begin{figure}[!ht]
  \begin{center}
    \includegraphics[height=2.3in]{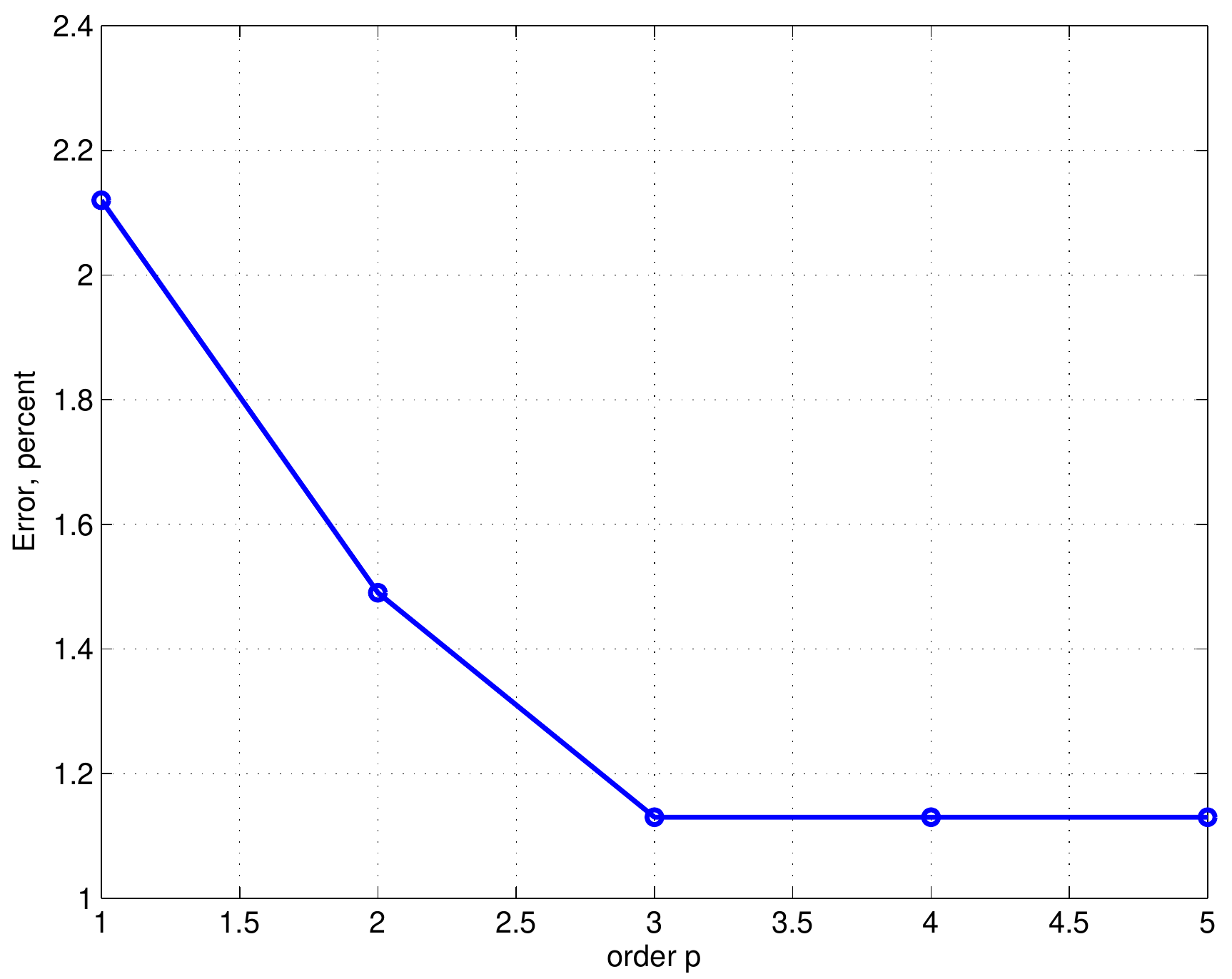}
  \caption{Total classification error as a function of $p$.}
  \label{res_p}
  \end{center}
\end{figure}

The parameters of the proposed approach include $p$, $q$, $K$, $\sigma_o^2$, and $\sigma_d^2$. Given the MNIST data set as an example, we further study the classification performance with different parameter values. Firstly, the performance for different $q$ values from $2$ to $40$ is measured, as shown in Fig. \ref{qerr}, when $K = 100$, $p = 3$, $\sigma_o^2 = 0.03$ and $\sigma_d^2 = 0.9$. At the far right is the performance of the Euclidean distance metric, which is equivalent to $q=N^2$. The performance is the best at $q=40$, the highest $q$ value tried in our experiments. Due to the slow performance improvement and the high storage requirement, we do not use a higher value. For $p=3$ and $q=40$, the performance as a function of $\gamma$ is shown in Fig. \ref{verr}, indicating that $\gamma=30$ is a reasonable value. Notice that our approach is the Euclidean distance metric classifier when $\gamma=1$. We also evaluate the performance as a function of $p$. The result is shown in Fig. \ref{res_p}, which shows no advantage for $p$ larger than $3$. We note that it is still possible to achieve better parameter settings, and different classification tasks would have different optimal parameter settings. 

\subsection{Comparison and Combination with Discriminative Classifier}
\begin{figure}[!ht]
  \begin{center}
    \includegraphics[scale = 0.85]{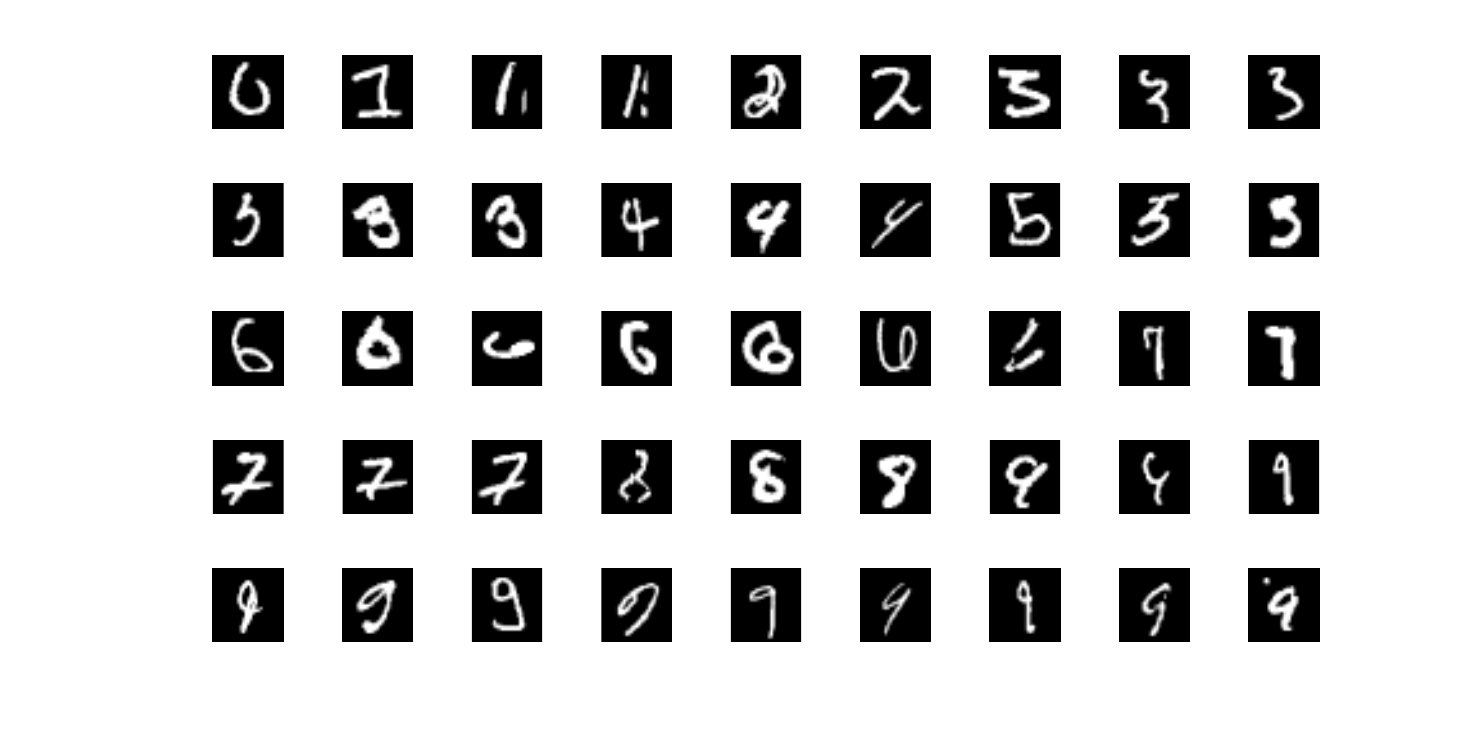}
  \caption{The total $45$ test images in the MNIST test data set that are correctly recognized by our Kernel-Distortion generative classifier but are misclassified by the $780 \times 500 \times 500 \times 2000 \times 10$ DBN.}
  \label{rbm_paul}
  \end{center}
\end{figure}

Our kernel-based generative classifier in the distortion space also provides a
diverse learning capability in character recognition compared with existing
discriminative classifiers. A well-trained DBN has a great
capability to learn complex models of feature representations with a much better recognition accuracy ($108$ out of $10,000$
images are misclassified for the MNIST data set). For the same MNIST data set, even
though our generative classifier has more misclassifications than the DBN, it is interesting to notice that ours can correctly classify, with high
likelihoods, $45$ out of these $108$ images that are misclassified in DBN,
demonstrating its very different recognition capability. We show these $45$
images in Fig. \ref{rbm_paul}, most of which just have some distortions from the training data. It also indicates that our proposed generative classifier has a nice
discriminative ability for these unseen test data that have some distortions
from the training data. 

Hybrid combination methods can be used to take advantage of these two different
types of learning algorithms. Both cascading and stacking hybrid methods were tested in our experiments. For the MNIST handwritten digits data set, compared to the discriminative classifier (DBN) and our proposed generative classifier which have a test error rate of $1.08\%$ and $2.38\%$, respectively, the cascading hybrid method
can decrease the test error rate to $1.04\%$ when the threshold $\tau$ equals 
$0.91$ and the stacking hybrid method can further decrease the minimum test error rate to $0.99\%$ when the weight $\hat{w}$ is $0.52$. Both $\tau$ and $\hat{w}$ are learned through 10-fold cross validation in the training data set. For the USPS handwritten digits data set, the test error rate is further reduced to $1.27\%$ using cascading hybrid method and to $1.00\%$ using stacking hybrid method from $1.45\%$ using our generative classifier and $3.09\%$ using DBN classifier. 

We also compare the classification performance of hybrid classifiers using naive Bayes, GMM, GKDE, LKDE and our proposed Kernel-Distortion classifier as the generative classifier, when MLP neural network and deep belief net are used as the discriminative classifier, respectively. We show the classification performance of these hybrid classifiers on the MNIST data set in Table \ref{comparsion_MNIST} and on the USPS data set in Table \ref{comparsion_USPS}, in which the best one with the minimum classification error rate is highlighted. These results show that our proposed generative Kernel-Distortion classifier has the best performance compared to the other four generative classifiers when combining with discriminative classifiers. 

\begin{table}[!ht]
\centering
\renewcommand {\arraystretch}{1.5}
\caption{MNIST data set: performance comparison of hybrid classifiers using five generative classifiers, including naive Bayes, GMM, GKDE, LKDE, and our proposed Kernel-Distortion classifier, when MLP neural network and DBN perform as discriminative classifier, respectively. All results are shown in percentage. The best one in each column is highlighted with underlined \textbf{\underline{Bold}} value.}
\label{comparsion_MNIST}
\centering
\setlength{\tabcolsep}{4.0pt}
\resizebox{12cm}{!}{
{\footnotesize
\begin{threeparttable}
\begin{tabular}{c c c c c }
\toprule
\multirow{2}{*}{\textit{MNIST Data Set}} & \multicolumn{2}{c}{MLP Neural Network ($2.48\%$)} & \multicolumn{2}{c}{Deep Belief Net ($1.08\%$)} \\
\cline{2-3} \cline{4-5}
& Cascading & Stacking & Cascading & Stacking \\
\midrule
Naive Bayes ($13.73 \%$) & $2.48\%$ & $2.46\%$ & $1.08 \% $ & $1.07 \%$  \\
GMM ($2.58 \%$) & $1.90\%$  & $1.88\%$ & $1.05 \%$ & $1.02 \%$\\
GKDE ($6.05 \%$) & $2.38\%$  & $2.35\%$ & $1.08 \%$ & $1.06 \%$\\
LKDE ($7.24 \%$) & $2.45\%$  & $2.33\%$ & $1.07 \%$ & $1.08 \%$\\
Kernel-Distortion ($\underline{\textbf{2.38}}\%$) & $\textbf{\underline{1.65}}\%$ & $\underline{\textbf{1.56}}\%$ & $\underline{\textbf{1.04}} \% $ & $\underline{\textbf{0.99}} \%$ \\
\bottomrule
\end{tabular}
\end{threeparttable}}}
\end{table}

\begin{table}[!ht]
\centering
\renewcommand {\arraystretch}{1.5}
\caption{USPS data set: performance comparison of hybrid classifiers using five generative classifiers, including naive Bayes, GMM, GKDE, LKDE, and our proposed Kernel-Distortion classifier, when MLP neural network and DBN perform as discriminative classifier, respectively. All results are shown in percentage. The best one in each column is highlighted with underlined \textbf{\underline{Bold}} value.}
\label{comparsion_USPS}
\centering
\setlength{\tabcolsep}{4.0pt}
\resizebox{12cm}{!}{
{\footnotesize
\begin{threeparttable}
\begin{tabular}{c c c c c }
\toprule
\multirow{2}{*}{\textit{USPS Data Set}} & \multicolumn{2}{c}{MLP Neural Network ($2.91\%$)} & \multicolumn{2}{c}{Deep Belief Net ($3.09\%$)} \\
\cline{2-3} \cline{4-5}
& Cascading & Stacking & Cascading & Stacking \\
\midrule
Naive Bayes ($10.45 \%$) & $2.91\%$ & $2.73\%$ & $3.09 \% $ & $3.00 \%$  \\
GMM ($3.82 \%$) & $2.73\%$  & $2.27\%$ & $2.00 \%$ & $1.91 \%$\\
GKDE ($6.09 \%$) & $2.82\%$  & $2.91\%$ & $2.82 \%$ & $2.73 \%$\\
LKDE ($2.64 \%$) & $2.18\%$  & $2.64\%$ & $2.18 \%$ & $2.27 \%$\\
Kernel-Distortion ($\underline{\textbf{1.45}} \%$) & $\underline{\textbf{1.36}}\%$ & $\underline{\textbf{1.27}}\%$ & $\underline{\textbf{1.27}}\%$ & $\underline{\textbf{1.00}} \%$ \\
\bottomrule
\end{tabular}
\end{threeparttable}}}
\end{table}

\section{Conclusions}
In this paper, we presented a novel kernel-based generative classifier, named Kernel-Distortion classifier, which is defined in a distortion subspace using the polynomial series expansion. Through distortion subspace analysis, predictable distortions of kernels, including translation, expansion and rotation, can be computed with differential linear operators. By incorporating these predictable distortions into our kernel-based model, the Kernel-Distortion classifier is able to improve the prediction on those distorted test samples that are usually misclassified even for a well-trained deep belief network. In our Kernel-Distortion classifier, we further developed an iterative kernel selection method to select the optimal kernels for modeling the distortion distribution. For the best use of the diversity introduced by the new classifier, we implement two hybrid combination schemes, cascading and voting, to increase classification accuracy. The experimental results demonstrate the effectiveness of our proposed classifiers.

\section{Acknowledgments}
The authors are grateful to the anonymous reviewers for providing comments and suggestions that improved the quality of the paper. This research is supported in part by National Science Foundation under grant ECCS 1053717 and CCF 1439011, and the Army Research Office under grant W911NF-12-1-0378. Any opinions, findings, and conclusions or recommendations expressed in this material are those of the author(s) and do not necessarily reflect the views of the National Science Foundation and the Army Research Office.

\section*{References}

\bibliography{ocr}

\begin{thebibliography}{10}
\expandafter\ifx\csname url\endcsname\relax
  \def\url#1{\texttt{#1}}\fi
\expandafter\ifx\csname urlprefix\endcsname\relax\def\urlprefix{URL }\fi
\expandafter\ifx\csname href\endcsname\relax
  \def\href#1#2{#2} \def\path#1{#1}\fi

\bibitem{jarrett2009best}
K.~Jarrett, K.~Kavukcuoglu, M.~Ranzato, Y.~LeCun, What is the best multi-stage
  architecture for object recognition?, in: IEEE International Conference on
  Computer Vision, 2009, pp. 2146--2153.

\bibitem{kato1999handwritten}
N.~Kato, M.~Suzuki, S.~Omachi, H.~Aso, Y.~Nemoto, A handwritten character
  recognition system using directional element feature and asymmetric
  mahalanobis distance, IEEE Transactions on Pattern Analysis and Machine
  Intelligence 21~(3) (1999) 258--262.

\bibitem{liu2007normalization}
C.-L. Liu, Normalization-cooperated gradient feature extraction for handwritten
  character recognition, IEEE Transactions on Pattern Analysis and Machine
  Intelligence 29~(8) (2007) 1465--1469.

\bibitem{hinton2006fast}
G.~E. Hinton, S.~Osindero, Y.-W. Teh, A fast learning algorithm for deep belief
  nets, Neural computation 18~(7) (2006) 1527--1554.

\bibitem{cirecsan2011handwritten}
D.~C. Cire{\c{s}}an, U.~Meier, L.~M. Gambardella, J.~Schmidhuber, Handwritten
  digit recognition with a committee of deep neural nets on gpus, arXiv
  preprint arXiv:1103.4487.

\bibitem{cirecsan2011convolutional}
D.~C. Cire{\c{s}}an, U.~Meier, L.~M. Gambardella, J.~Schmidhuber, Convolutional
  neural network committees for handwritten character classification, in:
  International Conference on Document Analysis and Recognition, 2011, pp.
  1135--1139.

\bibitem{cirecsan2011flexible}
D.~C. Cire{\c{s}}an, U.~Meier, J.~Masci, L.~M. Gambardella, J.~Schmidhuber,
  Flexible, high performance convolutional neural networks for image
  classification, in: Proceedings of the Twenty-Second international joint
  conference on Artificial Intelligence-Volume Volume Two, 2011, pp.
  1237--1242.

\bibitem{pan2010survey}
S.~J. Pan, Q.~Yang, A survey on transfer learning, IEEE Transactions on
  Knowledge and Data Engineering 22~(10) (2010) 1345--1359.

\bibitem{huang2012boosting}
P.~Huang, G.~Wang, S.~Qin, Boosting for transfer learning from multiple data
  sources, Pattern Recognition Letters 33~(5) (2012) 568--579.

\bibitem{wang2014unsupervised}
Q.-F. Wang, F.~Yin, C.-L. Liu, Unsupervised language model adaptation for
  handwritten chinese text recognition, Pattern Recognition 47~(3) (2014)
  1202--1216.

\bibitem{BagImDist}
P.~Baggenstoss, Image distortion analysis using polynomial series expansion,
  IEEE Transactions on Pattern Analysis and Machine Intelligence, 26~(11)
  (2004) 1438--1451.

\bibitem{duda2012pattern}
R.~O. Duda, P.~E. Hart, D.~G. Stork, Pattern classification, John Wiley \&
  Sons, 2012.

\bibitem{kay2016probability}
S.~Kay, Q.~Ding, B.~Tang, H.~He, Probability density function estimation using
  the {EEF} with application to subset/feature selection, IEEE Transactions on
  Signal Processing 64~(3) (2016) 641--651.

\bibitem{tang2016EEF}
B.~Tang, S.~Kay, H.~He, P.~M. Baggenstoss, {EEF}: Exponentially embedded
  families with class-specific features for classification, IEEE Signal
  Processing Letters 23~(7) (2016) 969--973.

\bibitem{tang2016toward}
B.~Tang, S.~Kay, H.~He, Toward optimal feature selection in naive bayes for
  text categorization, arXiv preprint arXiv:1602.02850.

\bibitem{tang2016bayesian}
B.~Tang, H.~He, P.~M. Baggenstoss, S.~Kay, A {Bayesian} classification approach
  using class-specific features for text categorization, IEEE Transactions on
  Knowledge and Data Engineering 28~(6) (2016) 1602--1606.

\bibitem{rosenblatt1956remarks}
M.~Rosenblatt, et~al., Remarks on some nonparametric estimates of a density
  function, The Annals of Mathematical Statistics 27~(3) (1956) 832--837.

\bibitem{parzen1962estimation}
E.~Parzen, On estimation of a probability density function and mode, The Annals
  of Mathematical Statistics (1962) 1065--1076.

\bibitem{scott2009multivariate}
D.~W. Scott, Multivariate density estimation: theory, practice, and
  visualization, Vol. 383, John Wiley \& Sons, 2009.

\bibitem{simonoff1996smoothing}
J.~S. Simonoff, Smoothing methods in statistics, Springer Science \& Business
  Media, 1996.

\bibitem{xie1993vector}
Q.~Xie, C.~A. Laszlo, R.~K. Ward, Vector quantization technique for
  nonparametric classifier design, IEEE Transactions on Pattern Analysis and
  Machine Intelligence 15~(12) (1993) 1326--1330.

\bibitem{tang2015parametric}
B.~Tang, H.~He, Q.~Ding, S.~Kay, A parametric classification rule based on the
  exponentially embedded family, IEEE Transactions on Neural Networks and
  Learning Systems 26~(2) (2015) 367--377.

\bibitem{memisevic2012shared}
R.~Memisevic, L.~Sigal, D.~J. Fleet, Shared kernel information embedding for
  discriminative inference, IEEE Transactions on Pattern Analysis and Machine
  Intelligence 34~(4) (2012) 778--790.

\bibitem{patwardhan2008robust}
K.~A. Patwardhan, G.~Sapiro, V.~Morellas, Robust foreground detection in video
  using pixel layers, IEEE Transactions on Pattern Analysis and Machine
  Intelligence 30~(4) (2008) 746--751.

\bibitem{huang2008metamorphs}
X.~Huang, D.~N. Metaxas, Metamorphs: deformable shape and appearance models,
  IEEE Transactions on Pattern Analysis and Machine Intelligence 30~(8) (2008)
  1444--1459.

\bibitem{comaniciu2003kernel}
D.~Comaniciu, V.~Ramesh, P.~Meer, Kernel-based object tracking, IEEE
  Transactions on Pattern Analysis and Machine Intelligence 25~(5) (2003)
  564--577.

\bibitem{marron1994transformations}
J.~S. Marron, D.~Ruppert, Transformations to reduce boundary bias in kernel
  density estimation, Journal of the Royal Statistical Society. Series B
  (Methodological) (1994) 653--671.

\bibitem{botev2010kernel}
Z.~Botev, J.~Grotowski, D.~Kroese, et~al., Kernel density estimation via
  diffusion, The Annals of Statistics 38~(5) (2010) 2916--2957.

\bibitem{rubinstein1997discriminative}
Y.~D. Rubinstein, T.~Hastie, et~al., Discriminative vs informative learning,
  in: In Proceedings of Third International Conference on Knowledge Discovery
  and Data Mining, Vol.~5, 1997, pp. 49--53.

\bibitem{nigam1999using}
K.~Nigam, J.~Lafferty, A.~McCallum, Using maximum entropy for text
  classification, in: IJCAI-99 workshop on machine learning for information
  filtering, Vol.~1, 1999, pp. 61--67.

\bibitem{vapnik1998statistical}
V.~Vapnik, Statistical learning theory (1998).

\bibitem{tang2015enn}
B.~Tang, H.~He, {ENN}: Extended nearest neighbor method for pattern recognition
  [research frontier], IEEE Computational Intelligence Magazine 10~(3) (2015)
  52--60.

\bibitem{drummond2006discriminative}
C.~Drummond, Discriminative vs. generative classifiers for cost sensitive
  learning, in: Proceedings of the Nineteenth Canadian Conference on Artificial
  Intelligence, Lecture Notes in Artificial Intelligence, 2006, pp. 479--490.

\bibitem{Mitchell97}
T.~Mitchell, Machine Learning, CHAPTER 1. GENERATIVE AND DISCRIMINATIVE.
  CLASSIFIERS, McGraw Hill, New York, 1997.

\bibitem{Long07}
P.~Long, R.~Servedio, H.~U. Simon., Discriminative learning can succeed where
  generative learning fails., Information Processing Letters 103~(4) (2007)
  131--135.

\bibitem{Shai2001}
S.~Fine, J.~Navratil, R.~Gopinath, A hybrid gmm/svm approach to speaker
  identification, in: International Conference on Acoustics, Speech, and Signal
  Processing, 2001.

\bibitem{ngdiscriminative}
A.~Y. Ng, M.~I. Jordan, On discriminative vs. generative classifiers: A
  comparison of logistic regression and naive bayes, in: Advances in neural
  information processing systems, no.~14, 2002.

\bibitem{jebara02discriminative}
T.~Jebara, Discriminative, generative and imitative learning, Ph.D. thesis,
  Massachusetts Institute of Technology (2002).

\bibitem{pernkopf2005}
F.~Pernkopf, J.~A. Bilmes, Discriminative versus generative parameter and
  structure learning of bayesian network classifiers., in: ICML'05, 2005, pp.
  657--664.

\bibitem{Drummond05}
C.~Drummond, Discriminative vs. generative classifiers: An in-depth
  experimental comparison using cost curves, nrc/erb-1135 nrc 48480, Tech.
  rep., NRC (Dec 2005).

\bibitem{Yuret2008}
M.~A.~Y. Deniz~Yuret, A.~E. Ural, Discriminative vs. generative approaches in
  semantic role labeling., in: Conference on Computational Natural Language
  Learning, 2008.

\bibitem{Schmah08RBM}
T.~Schmah, G.~Hinton, R.~Zemel, S.~Small, S.~Strother, generative versus
  discriminative training of rbms for classification of fmri images, in:
  Advances in Neural Information Processing Systems, 2008.

\bibitem{raina2003classification}
R.~Raina, Y.~Shen, A.~Y. Ng, A.~McCallum, Classification with hybrid
  generative/discriminative models., in: Proceedings of Neural Information
  Processing Systems, Vol.~16, 2003.

\bibitem{li2005generative}
Y.~Li, L.~G. Shapiro, J.~A. Bilmes, A generative/discriminative learning
  algorithm for image classification, in: IEEE International Conference on
  Computer Vision, Vol.~2, 2005, pp. 1605--1612.

\bibitem{fujino2005hybrid}
A.~Fujino, N.~Ueda, K.~Saito, A hybrid generative/discriminative approach to
  semi-supervised classifier design, in: Proceedings of the National Conference
  on Artificial Intelligence, Vol.~20, 2005, p. 764.

\bibitem{HolubWP08}
A.~Holub, M.~Welling, P.~Perona, Hybrid generative-discriminative visual
  categorization, International Journal of Computer Vision 77~(1-3) (2008)
  239--258.

\bibitem{btang2014hybrid}
B.~Tang, Q.~Ding, H.~He, S.~Kay, Hybrid classification with partial models, in:
  International Joint Conference on Neural Network, IEEE World Congress on
  Computational Intelligence, 2014.

\bibitem{bosch2008scene}
A.~Bosch, A.~Zisserman, X.~Muoz, Scene classification using a hybrid
  generative/discriminative approach, IEEE Transactions on Pattern Analysis and
  Machine Intelligence 30~(4) (2008) 712--727.

\bibitem{jaakkola1999exploiting}
T.~Jaakkola, D.~Haussler, et~al., Exploiting generative models in
  discriminative classifiers, Advances in neural information processing systems
  (1999) 487--493.

\bibitem{specht1990probabilistic}
D.~F. Specht, Probabilistic neural networks, Neural Networks 3~(1) (1990)
  109--118.

\bibitem{alpaydin1998cascading}
E.~Alpaydin, C.~Kaynak, Cascading classifiers, Kybernetika 34~(4) (1998)
  369--374.

\bibitem{alimoglu2001combining}
F.~Alimoglu, E.~Alpaydin, Combining multiple representations for pen-based
  handwritten digit recognition, Turkish Journal of Electrical Engineering and
  Computer Sciences 9~(1).

\bibitem{wolpert1992stacked}
D.~H. Wolpert, Stacked generalization, Neural networks 5~(2) (1992) 241--259.

\bibitem{hoeting1999bayesian}
J.~A. Hoeting, D.~Madigan, A.~E. Raftery, C.~T. Volinsky, Bayesian model
  averaging: a tutorial, Statistical science (1999) 382--401.

\bibitem{lecun1998gradient}
Y.~LeCun, L.~Bottou, Y.~Bengio, P.~Haffner, Gradient-based learning applied to
  document recognition, Proceedings of the IEEE 86~(11) (1998) 2278--2324.

\bibitem{hull1994database}
J.~J. Hull, A database for handwritten text recognition research, IEEE
  Transactions on Pattern Analysis and Machine Intelligence 16~(5) (1994)
  550--554.

\end{thebibliography}

\end{document}